\documentclass[msom,nonblindrev]{informs3}
\DoubleSpacedXI

\usepackage{natbib}
 \bibpunct[, ]{(}{)}{,}{a}{}{,}%
 \def\bibfont{\small}%

\usepackage{amssymb}
\usepackage{amsfonts}
\usepackage{booktabs}
\usepackage{multirow}
\usepackage{array}
\TheoremsNumberedThrough     % Preferred (Theorem 1, Lemma 1, Theorem 2)
\EquationsNumberedThrough    % Default: (1), (2), ...
\begin{document}
%%%%%%%%%%%%%%%%

% Outcomment only when entries are known. Otherwise leave as is and 
%   default values will be used.
%\setcounter{page}{1}
%\VOLUME{00}%
%\NO{0}%
%\MONTH{Xxxxx}% (month or a similar seasonal id)
%\YEAR{0000}% e.g., 2005
%\FIRSTPAGE{000}%
%\LASTPAGE{000}%
%\SHORTYEAR{00}% shortened year (two-digit)
%\ISSUE{0000} %
%\LONGFIRSTPAGE{0001} %
%\DOI{10.1287/xxxx.0000.0000}%

% Author's names for the running heads
% Sample depending on the number of authors;
% \RUNAUTHOR{Jones}
% \RUNAUTHOR{Jones and Wilson}
\RUNAUTHOR{Bertsimas et al.}
% \RUNAUTHOR{Jones et al.} % for four or more authors
% Enter authors following the given pattern:
%\RUNAUTHOR{}

% Title or shortened title suitable for running heads. Sample:
\RUNTITLE{Early Warning Index}
% Enter the (shortened) title:
%\RUNTITLE{}

% Full title. Sample:
\TITLE{Early Warning Index for Patient Deteriorations in Hospitals}
% Enter the full title:
%\TITLE{}

% Block of authors and their affiliations starts here:
% NOTE: Authors with same affiliation, if the order of authors allows, 
%   should be entered in ONE field, separated by a comma. 
%   \EMAIL field can be repeated if more than one author
\ARTICLEAUTHORS{%
\AUTHOR{Dimitris Bertsimas}
\AFF{Sloan School of Management, Massachusetts Institute of Technology, \EMAIL{dbertsim@mit.edu}, \URL{}}
\AUTHOR{Yu Ma}
\AFF{Operations and Information Management, University of Wisconsin Madison, \EMAIL{yu.ma@wisc.edu}, \URL{}}

\AUTHOR{Kimberly Villalobos Carballo}
\AFF{Technology Management and Innovation, New York University, \EMAIL{kv2361@nyu.edu}, \URL{}}

\AUTHOR{Gagan Singh}
\AFF{Hartford HealthCare, 
\EMAIL{gagan.singh@hhchealth.org}, \URL{}}

\AUTHOR{Michal Laskowski}
\AFF{Holistic Hospital Optimization, 
\EMAIL{michal.laskowski@hhopt.com}, \URL{}}

\AUTHOR{Jeff Mather}
\AFF{Hartford HealthCare, 
\EMAIL{jeff.matherh@hhchealth.org}, \URL{}}

\AUTHOR{Dan Kombert}
\AFF{Hartford HealthCare, 
\EMAIL{daniel.kombert@hhchealth.org}, \URL{}}

\AUTHOR{Howard Haronian}
\AFF{Hartford HealthCare, 
\EMAIL{howard.haronian@hhchealth.org}, \URL{}}

% Enter all authors
} % end of the block

\ABSTRACT{%
\textbf{Problem definition:} Capacity-constrained hospitals lack automated systems to harness the growing volume of heterogeneous clinical and operational data—ranging from vital signs and lab results to scheduling and patient flow metrics—to effectively forecast critical events. Early identification of patients at risk for deterioration is essential not only for patient care quality monitoring but also for physician care management. However, translating varied data streams into accurate and interpretable risk assessments poses significant challenges due to inconsistent data formats. \\
\textbf{Methodology:} We develop a multimodal machine learning framework, the Early Warning Index (EWI), to predict the aggregate risk of ICU admission, emergency response team dispatch, and mortality. Key to EWI’s design is a human-in-the-loop process: clinicians help determine alert thresholds and interpret model outputs, which are enhanced by explainable outputs using Shapley Additive exPlanations (SHAP) to highlight clinical and operational factors (e.g., scheduled surgeries, ward census) driving each patient’s risk. We deploy EWI in a hospital dashboard that stratifies patients into three risk tiers.  \\
\textbf{Results and Managerial implications:} Using a dataset of 18,633 unique patients at a large U.S. hospital, our approach automatically extracts features from both structured and unstructured electronic health record (EHR) data and achieves C-statistics of 0.796. It is currently used as a triage tool for proactively managing at-risk patients. The proposed approach saves physicians valuable time by automatically sorting patients of varying risk levels, allowing them to concentrate on patient care rather than sifting through complex EHR data. By further pinpointing specific risk drivers, the proposed model provides data-informed adjustments to care giver scheduling and allocation of critical resources (i.e. ICU beds). As a result, clinicians and administrators can avert downstream complications including costly procedures or high readmission rates and improve overall patient flow. 
}%

% Sample
%\KEYWORDS{deterministic inventory theory; infinite linear programming duality; 
%  existence of optimal policies; semi-Markov decision process; cyclic schedule}

% Fill in data. If unknown, outcomment the field
% \HISTORY{}

\maketitle
%%%%%%%%%%%%%%%%%%%%%%%%%%%%%%%%%%%%%%%%%%%%%%%%%%%%%%%%%%%%%%%%%%%%%%

% Samples of sectioning (and labeling) in MSOM
% NOTE: (1) \section and \subsection do NOT end with a period
%       (2) \subsubsection and lower need end punctuation
%       (3) capitalization is as shown (title style).
%
%\section{Introduction.}\label{intro} %%1.
%\subsection{Duality and the Classical EOQ Problem.}\label{class-EOQ} %% 1.1.
%\subsection{Outline.}\label{outline1} %% 1.2.
%\subsubsection{Cyclic Schedules for the General Deterministic SMDP.}
%  \label{cyclic-schedules} %% 1.2.1
%\section{Problem Description.}\label{problemdescription} %% 2.

% Text of your paper here

\section{Introduction}\label{intro}
Healthcare operations are increasingly relying on data-driven methods to forecast important logistical and clinical events. The forecast of deteriorating patients in particular is of significant financial and operational importance across hospital systems. A successful alert algorithm can help physicians proactively treat warning signs and apply immediate urgent care to prevent patients from further severe deterioration (\cite{Alam2014}), which reduces potential downstream costs of additional treatments, as well as increased risk of future readmission. These accurate planning is also instrumental in resource allocation, where successful predictions efficiently ensure specialized personnel being deployed and allow other staff members to stay focused on routine care, whereas unsuccessful predictions draw resources away from other deteriorating conditions and emergencies. From physicians’ perspective, advanced notice of ad-hoc deterioration events could also help reduce unpredictability in their workspace, which is a leading cause of their burnouts. However, in practice, such systems have been largely supported by front-line healthcare workers by their qualitative expertise using a limited subset of core patient information (\cite{Jones2006}), and their time can be significantly saved by using an automated, quantitative risk score prediction system. 

Our study setting is a large hospital system in the U.S., which has recently started to embed predictive models into its electronic healthcare record system. However, existing scoring systems, including the EPIC deterioration index (\cite{Byrd2023}), as well as NEWS scores (\cite{Smith2019}), were not adopted widely in practice by physician and nurse teams. The first predominant issue is their relatively low performance, caused usually by relying on only limited source of information from EHR and are thus not particularly personalized for individual patient profiles. These lead to frequent cases where predictions contradict clinicians’ medical judgments, which are then avoided altogether. In this study, we develop a multimodal learning framework to provide an accurate deterioration forecast called Early Warning Index (EWI). EWI integrates rich patient medical history and dynamically updates medical signals into a holistic framework to update deterioration risks daily to provide guidelines for physicians. Specifically, we jointly predict emergency response team dispatch, intensive care unit (ICU) admission, and near-term mortality in the next 24 hours as an aggregated deterioration event. Secondly, development of such deterioration systems in high-stake environments such as hospitals should not be purely quantitative. Rather, important operational and clinical insights, and private information that are not observable from EHR data, but equally significant in practice, should be integrated to iteratively refine model developments, and ultimately incorporated as part of the final decision. EWI facilitates these human-in-the-loop efforts through explainable feature quantification using SHAP, which computes the contribution of each input feature on the outcome and reveals important operational insights. To the best of our knowledge, there is limited work on leveraging this diverse level of data sources and human judgements into a single, holistic framework for patient deterioration risk prediction that has been implemented in practice. 

\subsection{Related Literature}
\subsubsection{Patient Deterioration Prediction}
\hfill

\noindent
Advanced notice of patient deterioration is closely tied to several important operational planning and forecast problems such as nurse scheduling (\cite{Legrain2024}), ICU bed management (\cite{Ouyang2020}), and future readmission (\cite{Mann2024}). Once a patient is admitted into the hospital, multiple caregiver teams will be mobilized to optimize the patient’s journey within the hospital. Importantly, when a patient is in critical condition or in need of emergency responses, significantly more resources will be allocated for their continuous monitoring and related medical needs. These involve re-appointing nurse managers or cardiac/respiratory specialists from their existing work shift to frequently check upon deteriorated patients, as well as potentially transferring patients to ICU units or step-down units. Unfortunately, both emergency-expertise clinicians and available beds are scarce and cost-intensive resources in most healthcare settings. It is thus of high importance for the upstream departments, including emergence departments and telemetry units, to have an advanced prediction of a potential need for a deterioration event occurrence and for the downstream departments to prepare for the receiving of a patient who undergo these events. This advanced knowledge of patients’ risk of deterioration can assist medical administrators in allocating this critical resource (\cite{KC2012, Bertsimas2022}) optimally. If not managed appropriately, resulting in flow congestions and over-capacitation could delay patients’ access and quality of care ( \cite{Bertsimas2022, Kim2015, Chen2023}), increase spillovers across other hospital units (\cite{Kim2024}), and even increase readmission risks if some patients are forced to be either discharged prematurely or be exposed to infection risks due to prolonged stay (\cite{Chan2012}). 

There are two types of deterioration events in clinical settings: acute and stable. Acute deterioration events are urgent, arising within high-stakes environments that demand rapid assessments of extensive patient data and immediate action, such as resuscitation. For instance, emergency response teams typically have less than five minutes to react following a dispatch call (\cite{Jung2016}). Current systems for detecting acute deterioration predominantly employ a "track-and-trigger" method. This involves continuous bedside monitoring that prompts immediate intervention upon recognizing critical abnormalities through a rule-based system (\cite{Gao2007}). Such systems force caregivers to react in real-time without the opportunity for advanced planning. On the contrary, stable deterioration refers to a more gradual, ongoing decline in a patient’s health, where physicians can typically anticipate and strategize interventions more. 

To this end, automated machine learning models have been proposed to stratify patients into different risk groups to assist physicians’ decision-making (\cite{Gao2020, Taylor2016}). These early warning alert systems include the National Early Warning Score (NEWS) (\cite{Smith2013}), the APACHE III prognosis system (\cite{Knaus1991}), as well as the EPIC deterioration index (\cite{Byrd2023}). However, variables included in the model development phase are often constrained to a single modality, or a source of data information.

\subsubsection{Multimodal and LLM Patient Representation}
\hfill

\noindent
The integration of data from diverse data modalities was one of the main sources for recent artificial intelligence’s superior performance across medical domains. During routine rounds of patient status updates, nurses and doctors are often given large quantities of data ranging from medications, procedures, and lab results to diagnoses for multiple patients. The difficulty to accurately integrate all this information manually in a short time spang makes room for error and could cause physicians fail to prioritize patients requiring the most immediate attention (Na 2023). Frameworks of multimodal learning have been proposed in several works (\cite{Chen2024, Acosta2022}) to unify data from tabular, time-series, vision, and language. However, despite the richness of information these novel methods provide, available predictive models still largely rely on a subset of limited medical information, often predominantly including age, gender, past medical diagnoses, and family history (\cite{Byrd2023, Smith2019}). The bottleneck of pushing for more personalized decision-making calls for need of an automated data processing pipeline that can scale across hospitals. However, processing different modalities of data into a consistent representation is not trivial. Challenges in missing data (\cite{Norris2000}), differences in patient record length, differences in service availabilities in different institutions (healthcare disparity) (\cite{Soto2013}), heterogeneity of data capture (\cite{Hilton2018}), recording errors (\cite{Bell2020}), and administrative input delays (\cite{Fadlullah2018}), pose bottlenecks for decisions to be made in real-time, high-stake environments.

Another large volume of literature is dedicated to the learning of textual data, which led to the recent breakthroughs in Large Language Models (LLMs). These versatile models can be applied to a wide range of tasks. An especially useful usage of these models for healthcare data is the representation of categorical information, which is traditionally processed via methods such as one-hot encoding or top-k frequency (including only the first k categories ranked by frequency). These traditional processing is sub-optimal especially in healthcare operational settings requiring continuous daily update: consider we wish to maintain medications categories prescribed to all patients, due to the sheer volume of possible categories, if all categories are included, this data matrix would be sparse. On the other hand, maintaining only a subset of frequently prescribed medications could lose important signals of patient condition. Instead, by considering the categorical information of EHR data as textual information, we effectively retain all raw signals. We discuss this more in detail in Section 2.3.

\subsubsection{Human-in-the-loop Decision Making}
\hfill

\noindent
With the increasing awareness of machine learning’s performance, a diverse pool of algorithms and models have been developed, piloted, and implemented across different operational decision areas. While these algorithms often exhibit impressive performance, there is an increasing recognition that emphasizes the important role of integrating human experts’ opinions, decisions, and insights to achieve practically relevant outcomes. In particular, such human-in-the-loop practices are shown to effectively complement and guide final operational and financial decisions across contexts including ride hailing (\cite{Benjaafar2024}), medication prescription (\cite{Baucum2023}), investment decisions (\cite{Bianchi2024}), hotel pricing (\cite{Garcia2024}), and text messaging (\cite{Teeni2023}). An overarching theme of this stream of literature is the recognition that humans often remain the ultimate decision makers, especially in high-stake settings. Therefore, machine learning algorithms should thus explore beyond algorithmic accuracy to examine and integrate how managers and other decision makers engage with model outputs based on real-world constraints such as resource limitations, human preferences, and organizational policies. In our specific setting, we build on these findings by incorporating human (i.e., caregiver and logistic managers) expertise in two phases. First, during model development, experts help identify negative corner cases—situations where the model fail due to data limitations or unrecognized medical conditions. Second, at deployment, they refine alert thresholds to account for organizational constraints such as limited ICU bed capacity and the risk of alert fatigue. We detail these human-in-the-loop design considerations and their implications for operational performance in Section 3.3.

\subsection{Contributions and Structure}
We propose a hospital-centric methodology that integrates historical medical and operational information of a patient’s admission stay to forecast near-term future deterioration risk. Our method involves two key steps that had not been previously explored or discussed: (a) an end-to-end machine learning pipeline that integrate multimodal data from a diverse source of information, (b) complementing quantitative evaluation of model performance with human-in-the-loop evaluations and model refinement. 

For the first step, we developed an automated extraction method to gather relevant tabular, time-series and language data from an existing large scale electronic health record system from a large U.S. hospital system. An important contribution of our work is tackling traditional challenges associated with integrating data of different formats and occurrence intervals. For example, we extensively discuss why we developed large language model-based method to reformulate medication and diagnosis categories as textual data to avoid sparsifying feature representation and lose rare occurred, but critical medical information. Instead, our approach exploits known advantages of LLM’s ability to gather contextual information as well as tapping into the larger store of knowledge it was originally pretrained on (see Section 2.3).

The second key contribution of our work is the integration of human-in-the-loop expertise throughout the model development, evaluation, and implementation process. Compare to previous literature which largely focuses on improving quantitative performance of similar forecast models, our approach integrates physician feedback as a key metric of success. We discuss extensively and provide concrete cases where such insights provided guidance on feature and model selection that comply with both existing medical understanding and operational constraints (see Section 3.3) and discuss in depth how human decisions guided the design of the developed alert system. This stream of analysis was made possible with explainability method that visualizes the contribution of features to each prediction, which quantitatively provided insights into relevant operational factors (i.e. service load within ward), administrative conditions (i.e. day of week), and clinical factors (i.e. vital measurements). 

The rest of the paper is organized as follows: In Section 2, we define the problem setting, present the patient cohort and the definitions of the deterioration outcomes, describe the multimodal patient representation. In Section 3, we outline our model training and evaluation methodology and detail how to adopt human-in-the-loop expertise. In Section 4, we report and compare the predictive power of multimodal combinations using three machine learning techniques on the retrospective data and discuss the operational insights we observed in the best-performing model. In Section 5, we outline the practical implementation pipeline and considerations and showcase the implemented dashboard. In Section 6, we discuss some managerial implications and limitations of the work and its impacts on operational efficiencies and patient outcomes.

\section{Deterioration Forecast Model}
\subsection{Problem Definition}
Our objective is to continuously predict near-term patient deterioration risk within their hospital admission stay using EHR data. We assume each patient’s medical information is indexed by \(t\), with \(x_t\) denoting the collection of all information that is available to us from admission to time \(t\). This information consists of three different sources of information, or modalities: tabular (denoted as \(tab_t\)), time-series (denoted as \(ts_t\)) and textual (denoted as \(text_t\)) and can be represented as \(x_t=[tab_t, ts_t, text_t]\). We describe in detail how to process each modality in Section 2.3 and 2.4. The granularity of t can adjust in practice depending on the use case, and our problem formulation is agnostic to its specific value. We adopt \(t\) as a single day (24 hours) as this interval collects ample information update in the participating hospital and announce these predictions at 8am in the morning to most closely adhere to physicians’ rounding schedule. We suppose that there is a function \(\beta(x_t)=(\beta_1,\beta_2, \cdots ,\beta_n )_t\) that aggregates all learned information up until time t. We call \(\boldsymbol{\beta}(x_t)\) the patient embedding at time \(t\), and refer to it as \(\boldsymbol{\beta}_t\) in the remainder of the paper. 

Let \(\boldsymbol{y}\) denote the time-series of deterioration outcome (defined later in Section 2.5), where \(y_t\) denotes whether a deterioration outcome has occurred during time \(t\) and \(t+1\). Given input \(\boldsymbol{\beta}_t\), we look for the best performing predictive model \(\mathfrak{F}\) that outputs a probability of outcome \(y_t\). In our setting, this is equivalent to asking: given all gathered medical information up until the end of day \(t\) of a patient’s stay, what is the probability of them developing a deterioration event in the next 24 hours.  

Existing approaches adopt a continuously updating patient-centric time scale, where a prediction is made after a designated period, starting with the time of a patient’s admission. However, in practice, daily rounds of physicians and nurses for the discussions of patient conditions are often caregiver-centric, where a designated time (usually twice a day) is used. We thus aggregate patient information ending at 11:59 pm every day and make predictions with this sharp deadline.

\subsection{Patient Cohort Selection}
A total of 18,633 patients were retrospectively identified from the existing hospital database from 2021-01-01 to 2023-04-01. We only include patients who are currently not in ICU units, as these patients are already in critically ill conditions, and life-threatening events such as emergency response team dispatch or mortality will be managed distinctively from other departments.  Patients who did not stay at the hospital overnight (length of stay of fewer than 24 hours), have an abnormal deterioration event dispatch time that is outside of their admission time window, and have missing admission dates are excluded from the patient cohort as well. Our cohort design aims to dynamically update patient risks daily throughout the admission, and thus, we consider 110,900 patient-admission-days, with each sample aggregating all previous days’ information of a patient admission into a single representation of \(\boldsymbol{\beta}_t\).  

\subsection{Textual Representation of Categories}
One bottleneck to utilizing medication and medical diagnoses data is the large number of categories within each source, where the majority of the categories only occur in a few patients. Traditional approaches take the most frequently occurring categories and discard the rest, thus leaving a large amount of information originally available unused. To address this challenge, we leverage recent advancements in Large Language Models (LLM) using pre-trained language models to represent time-series data (\cite{Wang2022, Huang2020, KC2012}). These methods project the original input text into a fixed-sized numerical vector representation, embedding, that can be used for downstream tasks. Concretely, given \(text_t\) as the input text that has been tokenized to \(m\) tokens, we use a pre-trained function \(f(\cdot,\theta)\), in this case, an LLM, to obtain a tensor of dimension m times a pre-defined embedding size. 
\(f(text_t,\theta): \mathbb{N}^m \rightarrow \mathbb{R}^{m \times \text{embedding size}}\)
This token can be further reduced to be a one-dimensional vector via averaging over the token dimension, effectively by:
\[g(f(text_t,\theta))_t=\frac{1}{m}\sum_{t=1}^m f(text_t,\theta)_t \]
The choice of the pre-trained LLM is crucial for accurate representation of the original language as well as computational efficiency for practical implementations. ClinicalBERT (\cite{Alsentzer2019}) is an open-source, state-of-the-art medical LLM built on the BERT architecture and finetuned on electronic health records from the MIMIC dataset (\cite{Johnson2016}). However, to avoid the computational hurdle of its relatively slower processing on a large amount of data, we instead use a distilled version of ClinicalBERT, tiny-ClinicalBERT (\cite{Rohanian2024}), that tries to mimic the teacher models’ behaviors with reduced computational costs and faster processing time. In addition to passing the original text information, we also incorporate contextual knowledge in the form of a prompt that informs the prediction task. The input reads, "\textit{We aim to predict if the patient will experience a deterioration event in the next 24 hours (emergency response team, ICU admission, mortality). [rest of category information].}" For each recorded patient-admission-day-time medication or diagnosis, we process the original text into these fixed-size embeddings of dimension 312.

\subsection{Multimodal Patient Representation}
A rich set of patient data from multiple sources and modalities is aggregated into a comprehensive patient profile. These features capture information from three main categories: tabular features, time-series features, and textual features. Specifically, tabular data include 1) demographic information (e.g., age); 2) social behavior (e.g., whether is an alcohol user); 3) patient administrative condition (e.g., whether the patient is in ICU), and 4) auxiliary operational variables which are not patient-specific (e.g., day of the week, ward utilization); time series data include: 5) vital signs (e.g., blood pressure, oxygen level), 6) lab results (e.g., potassium level); and textual data include 7) medication administered (e.g., medication name, dosage, and intake form), and 8) medical diagnosis (e.g., recorded by ICD codes). We process these features based on their native forms to optimize the extraction of their information. For static features, we encode categorical features so that they are ordered properly with respect to the outcome (i.e., alcohol user has a higher value than non-alcohol users). For time-series features whose precise numerical fluctuations signal patient conditions, we compute the daily average, peak, minimum, and last values recorded, as well as a trend from the previous day. For all features, when missing data is present, we first impute it by using the latest value recorded for that patient in the same hospital stay, and if it is missing for all admission days, we impute it with 0. 

The final set of features extracted was inspired by discussions with the physicians regarding certain medications prescribed to severe patients, which are summarized in Appendix Table 1. Two types of medications are considered: green and yellow. Specifically, green medications refer to intravenous medications that signify a patient’s relatively sick condition.  If these medications are discontinued, transitioned to oral medication, or no other medication from the same class was prescribed, this implies that the patient is improving. The inclusion of these medications’ prescription changes could thus be a signal of a non-inference event (not deteriorating).  Another set of yellow medications is generally used either during the procedure, in the ICU, or when the patients experience emergency responses. We assume that the use of these medications is a surrogate marker for sickness. These two groups of medications can also be used together to signify the patient's condition. For example, if a patient had a yellow medication during the hospitalization and is now on green medication, we can hypothesize that it could be an indicator that the patient is at higher risk for deterioration. 

To aggregate all information belonging to the same day, we further average all embeddings from the same day into a single patient-admission-day numerical vector. Encompassing all language embedding and other features, 1478 features were eventually extracted to provide a holistic representation of the entire patient profile. We remark that we do not adopt embedding alignment methods, such as contrastive learning, from these three different sources although it is possible that such approaches could improve final performance. A key to this decision is to ensure explainability at the model refinement stage to concretely identify the features and their respective raw values that had contributed to the prediction of deterioration risk, as is detailed in Section 3.3.

\subsection{Outcome Definitions}
We make predictions for every inpatient in the hospital for the following three deterioration events: an emergency response team dispatch, an ICU admission, or mortality in the next 24 hours. Specifically, an emergency response team dispatch consists of four types of events: rapid response team (RRT), cardiac alert team (CA), anesthesia STAT (AS), and difficult airway team (DART) dispatches. The majority of these calls come from emergency departments with neurology (mental status change, lethargic), cardiac (heart rate (HR)<40, heart rate>130, systolic blood pressure (SBP)<90, chest pain, new arrhythmia,), and respiratory reasons (oxygen saturation (SpO2) <90\%, respiratory distress, threatened airway). When an event is not recorded, we consider the occurrence to be negative (0). The aggregated deterioration is defined as the union of all three deterioration events, where the occurrence of any suffice to trigger a deterioration alert. Rigorously, this can be expressed as follows, where l is an indication function of the occurrence of an event: \[\mathbf{1}_{\text{deterioration}} =\mathbf{1}_{\text{emergency response}} +\mathbf{1}_{\text{ICU admission}} +\mathbf{1}_{\text{expiration}} .\]

We remark that these outcome variables are surrogate to our true objective: patient deterioration severity, and that the identifications of them are subjective to endogeneity. In particular, admission into the ICU unit could be determined beyond patient conditions, but instead are results from availability of ICU beds, as well as human decisions that are subject to uncertainties and biases. Our predictive model, thus, should also not be viewed as an estimate of how likely emergency response teams will be dispatched the next day, or whether or not the patient should be admitted into the ICU units. Instead, these machine learning models provide guidelines to estimate the likelihood of deteriorating patient conditions which could result in these critical events. 

\section{Model Development and Human-in-the-loop Refinement}
\subsection{Cohort Splitting}
Following traditional machine learning practice, the data set is split into training/validation/test sets, where training is used to develop models, validation is used to select the best model, and test is used to evaluate the final model performance on the unseen, out-of-sample dataset. To reflect and simulate real-world development cycles where the models are trained on past data and utilized for future predictions, data split is performed chronologically. We sort the data by record date and split it into 70\% training (77630 patient-admission-days), 15\% validation (16635 patient-admission-days), and 15\% testing sets (16635 patient-admission-days). Specifically, the training set includes all patient records from 2021-01-01 to 2022-08-07, the validation set includes all patients between 2022-08-13 and 2022-11-29, and the testing set includes all patients between 2022-11-29 and 2023-04-01. We note that this stratification by date and patient is to ensure that no data leakage is shared between these three groups. In comparison to the random splitting approach, under a practical production lens, having access to future dates’ samples and outcomes are inherently leaking information to the current training process.

\subsection{Model Training and Evaluation}
We apply three different types of machine learning models, random forest (\cite{Breiman2001}), gradient boosted trees (\cite{Chen2016}), and TabNet (\cite{Arik2021}), to learn the binary prediction task of two outcomes: occurrence of inference event (1) and non-event (0). The inference process aims to solve an empirical risk minimization (ERM) problem, where given \(T\) input realizations of past data with n features \(\beta_t \in \mathbb{R}^n, t=1,\cdots,T\) and their respective binary outcome variable \(y_t \in \{0,1\},t=1,\cdots,T\) we look for the optimal function solution \(\mathfrak{F}\) within a class of cleaners \(\mathcal{F}\) that minimizes the risk given a loss function 

\[\min_{\mathfrak{F} \in \mathcal{F}} \frac{1}{T} \sum_{t \in [1:T]} \ell(y_t, \mathfrak{F}(\boldsymbol{\beta}_t))\]

To reduce this infinite dimension optimization to a tractable form, we consider instead a parametric family of learners \(\mathfrak{F}_{\theta} (\cdot)\) with finite-dimension parameter \( \theta \). However, ERM approaches potentially lead to non-optimal out-of-sample performances, where a new sample \(\beta\) from the true data-generating distribution could potentially have a high loss due to optimal learners’ overfitting on the training data. To avoid this, we conduct regularization and hyperparameter tuning on the hold-one-out validation set to select the learner with the best generalizability performance. For tree-based models (random forest and gradient boosted trees), we select the optimal combination of the number of estimators (20, 50, 100) and maximum depth (3, 4);  for TabNet, we select the optimal combination of the dimension of the decision prediction layer (16, 32), dimension of the attention layer (16, 32), and number of steps in the decision process (3, 5). The optimal combination is used to conduct a final training on the combination of training and validation sets to optimize performance. 

We compute the accuracy metric of the area under the receiver operating curve (AUROC) reported on the test set and evaluate the sensitivity and specificity of the final learner to select the optimal threshold to modify the continuous probability output into a binary prediction. Specifically, sensitivity (recall) refers to the model's ability to predict a deterioration event when it occurs (captures the truly severe patients), and specificity refers to the model's ability to predict a non-deterioration event when it doesn’t occur (captures the non-severe patient). We also investigate precision (positive predicted value), which refers to the model’s accuracy of the positive predictions (predicted early warnings are indeed deteriorating patients).

\subsection{Human-in-the-loop Refinement}
In high-stake environments such as healthcare, directly implementing purely quantitative, machine-learned algorithm can lead to unintended negative consequences. Due to healthcare setting’s heavily human-oriented and resource-constrained nature, algorithms are at large developed as support tools complementing physicians’ decisions rather than autonomous decision makers. Care providers and logistic managers often have access to additional private information, as well as in-depth understanding of medical implications and the availability of existing resources to act upon these forecasts. In our setting, such expertise heavily guided model development in two phases: identifying negative corner cases, and refining alert thresholds. Once initial models are developed, we identify patients whose predictions significantly deviate from the ground truth, and use SHAP to quantify the most significant contributing factors that has resulted in the incorrect risk prediction. For example, a group of high-risk alerts were repeatedly generated for patients whom we later identified with DNR (do not resuscitate) orders. This highlights factors beyond medical conditions, in this case patient’s end-of-life preferences, were not effectively integrated in the system to prevent aggressive cares.  In another instance, certain medications were interpreted as signals of acute deterioration risk, though upon physician review, they were identified as care providers’ existing recognition of a patient’s declining condition. This led to the classification of green and yellow medications discussed previously. By iteratively removing, reclassifying, or adding these features into the predictive model, we align these algorithmic outputs with established existing clinical protocols and physician expectations more closely. 

Another important expert-driven decision of the forecast model is the setting of alert thresholds. Specifically, the raw output of the predictive model is a probability indicating the degree of patient’s likelihood of developing a deteriorating event. However, such raw output cannot be used directly in practice as they are not directly related to the decision or action needed by care providers. Instead, these probabilities are stratified into three groups: low, medium, and high risks, which then correspond to differing levels of alerts to notify them regarding who should be prioritized during their routine rounding check-ins. However, the identification of these thresholds is not trivial: if set too low, the predictive model will have high false negative rates and fail to alarm care providers of patients who deteriorate without being detected. This is a particularly negative scenario as such algorithmic failure could result in patient severe conditions and even mortality. On the other hand, it is important to recognize that hospitals are resource constrained environments, and the prevalent phenomenon of false positive alarms, where alerts go off for healthy patients, is a significant factor of physician burnout. These events divert resources away from patients who may truly need them and contribute to physicians’ reluctance to adopt these algorithms in the future, as they disrupt their work schedule. Thus, the decision of the thresholds of when to raise these alerts are extensively discussed with the physicians, where multiple metrics beyond performance are used to determine the final cutoffs. Among these, thresholds were implemented and tested in practice to gather physician feedback daily, and iteratively adjusted until a stable threshold was agreed upon among physicians who actively monitor the forecasts.

\subsection{Computation Resources}
Data processing, feature engineering, model training, and inference are conducted in Python 3.8.11 with a parallelization strategy on a remote server with 32GB RAM Intel Xeon-P8 CPU per instance. This implies that our framework is feasible from both compute and memory for majority of the hospital institutions. 

\section{Quantitative and Qualitative Performance}
\subsection{Accuracy of Model}
Among the three machine learning models experimented with, we obtained the highest test AUC using the gradient boosted trees model combining all three available modalities. In comparison to the best performing single modality model, the best performing multimodal model achieves a performance improvement of 4.7\%. The general trend of model performance improves with additional modalities included across different settings, with few exceptions where inclusion of additional modalities reduces performance (i.e. random forest tabular and language, in comparison to tabular data alone), which potentially implies that there are cases where increased number of features could imply more noise. 

% We also compare our alert system performance with the EPIC deterioration index, which is a golden standard used by many hospitals system that has EPIC’s electronic healthcare record system. In this cohort, we consider only test set patients who are both included in our cohort study and those who received an EPIC index in their record, which reduces the cohort to 10851 patients and 64190 patient-admission-days. We normalize the EPIC score, which was in a range of 0-100, to a probability range of 0-1 and consider this as the EPIC-proxy of early warning alert. We also note that EPIC scores are continuously updated multiple times across the entire day, and we consider the average probability obtained for a patient in a day. We obtain an AUC of 0.658 for EPIC and an AUC of 0.852 for our trained models, approximately a 29.5\% improvement in AUC performance. A detailed comparison of the model development difference between the EPIC deterioration index and our model can be found in Appendix Table 2. 

\begin{table}[h]
\centering
\fontsize{11}{7}
\begin{tabular}{@{}llccc@{}}
\toprule
\textbf{Data Combinations} & \textbf{} & \textbf{Random Forest} & \textbf{Gradient Boosted Trees} & \textbf{TabNet} \\ 
\midrule
\multicolumn{5}{l}{\textbf{Single Modality}} \\ 
\midrule
\multirow{1}{*}{ Tabular only}  &  & 0.738 & 0.760   & 0.668  \\ 
\multirow{1}{*}{ Time series only} & & 0.740 & 0.768  & 0.679 \\ 
\multirow{1}{*}{ Language only} & & 0.631 & 0.660 & 0.605 \\ 
\midrule
\multicolumn{5}{l}{\textbf{Two Modalities}} \\ 
\midrule
\multirow{1}{*}{ Tabular \& Time series} &  & 0.753  & 0.797  & 0.704 \\ 
\multirow{1}{*}{ Tabular \& Language}    
&  & 0.733  & 0.769    & 0.650  \\ 
\multirow{1}{*}{ Time series \& Language}& 
& 0.741   & 0.775   & 0.691 \\ 
\midrule
\multicolumn{5}{l}{\textbf{All Modalities}} \\ 
\midrule
\multirow{1}{*}{ Tabular \& Time series \& Language} &  & 0.754 & 0.798  & 0.708 \\ 
\bottomrule
\end{tabular}
\caption{Comparison of performances across different data combinations using Random Forest, Gradient Boosted Trees, and TabNet.}
\label{tab:comparison}
\end{table}

\subsection{Physician Determined Alert Threshold}
\begin{figure}[h!]
    \centering
    \includegraphics[width=\textwidth]{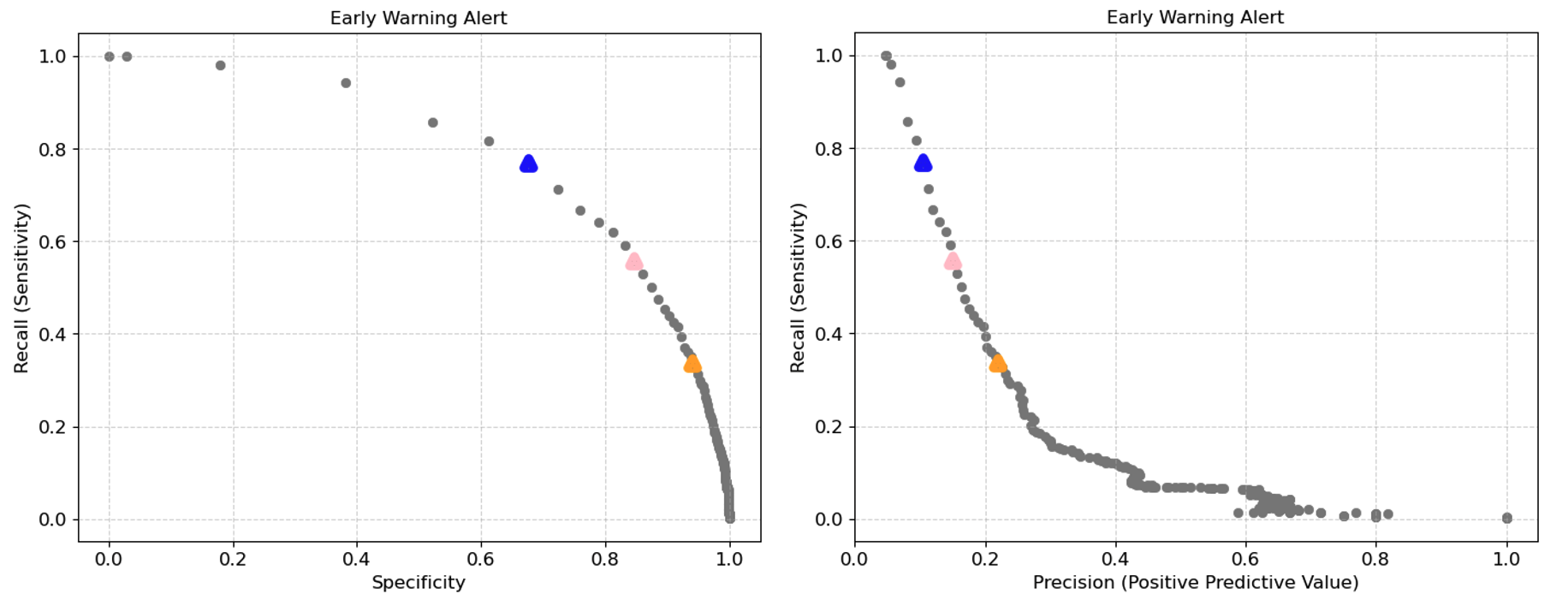} 
    \caption{Comparison between specificity and precision (positive predictive value) against recall (sensitivity) and the different options for choosing early warning alert threshold cutoffs. }
    \label{fig:fig1}
\end{figure}

In Figure \ref{fig:fig1}, we demonstrate the different options of probability threshold cutoffs of 0.03 (blue), 0.06 (pink), and 0.12 (orange), which lead to different sensitivity, specificity, and precision performances. The most optimal tradeoff emphasizes capturing all patients with a risk of deterioration, as a missed prediction could be life-threatening for the patient. Given a selected threshold (i.e., colored pink) has a sensitivity of 0.846, a specificity of 0.557, and a precision of 0.154 and is used as the cutoff for the early warning alert (all predictions higher than 0.06 is predicted as an alert, otherwise as not an alert). This alert will then be sent to the physicians’ team to request additional attention for the predicted medium-risk patients. Similarly, high-risk alerts can also be sent once they had reached a threshold of 0.12 (colored orange) given the tradeoffs from below. 

\subsection{Explainability by SHAP}
The ability to understand qualitative insights generated by machine learning models guide physicians to make better real-world medical decisions. SHAP values are considered a state-of-the-art interpretability method for quantifying feature-level contribution to predictions from black-box models. Specifically, a positive SHAP value implies that the inclusion of the feature contributes to predicting the occurrence of a deterioration event, and a negative SHAP value implies a contribution to predicting the non-occurrence of the deterioration event. 

\begin{figure}[h!]
    \centering
    \includegraphics[width=0.8\textwidth]{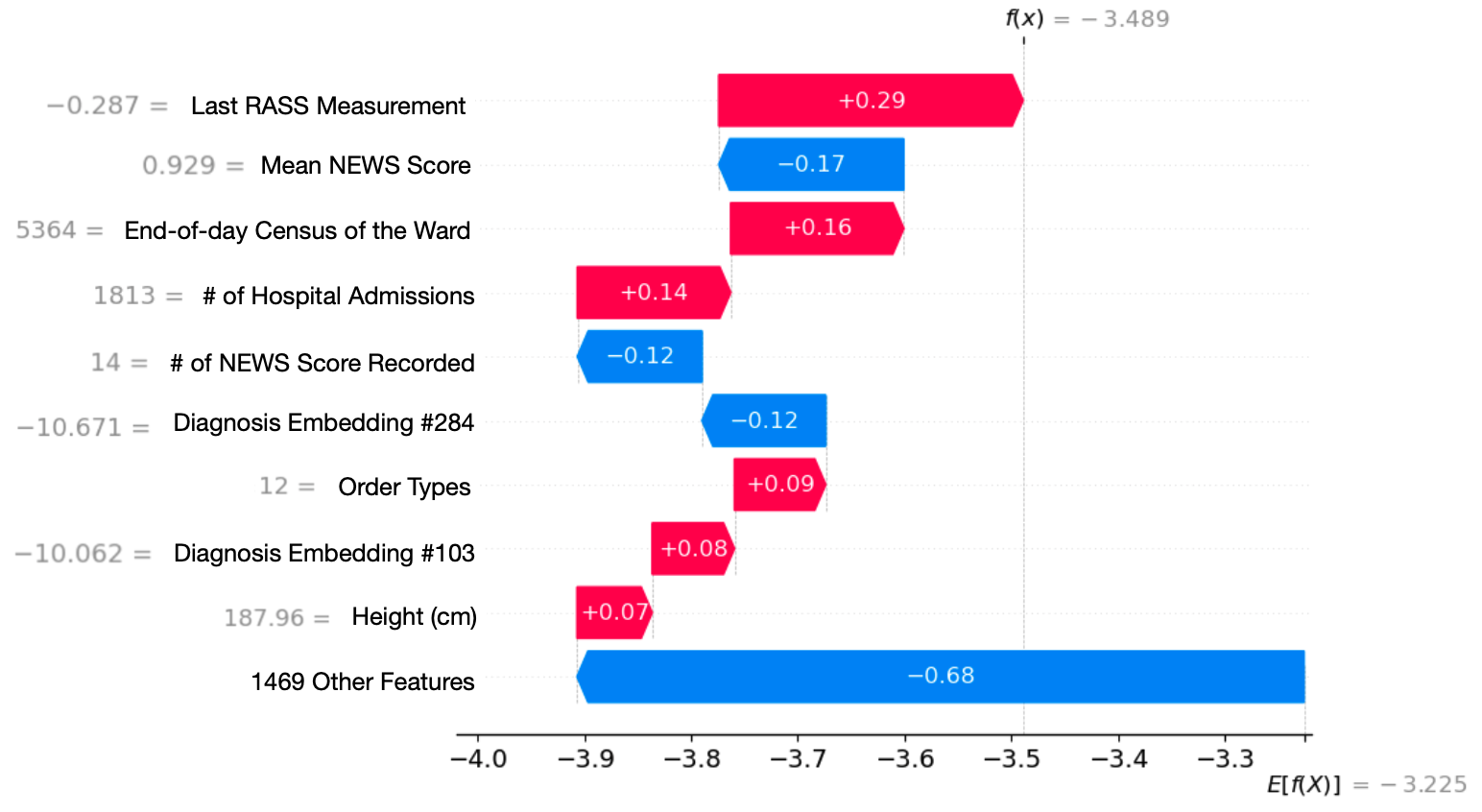} 
    \caption{SHAP plot for a single patient profile and demonstration of how different clinical and operational features contribute to the prediction of deterioration risk.   }
    \label{fig:fig2}
\end{figure}

Consistent with previous operations management literature, observations from SHAP indicate that the prediction of deterioration events is strongly correlated with operational factors. In the example indicated for an individual patient, a higher end-of-day census of the ward, which corresponds to a more congested ward facility, intuitively reduces individual-level care quality. Similarly, a high number of hospital admissions could cause potential reduced attention of care providers to individual patients due to high volume of traffic across the entire hospital facility. These factors indicate that deterioration events are far from simple demonstrations of medical complications, but rather also a combination of with hospital service quality as well. 

\begin{figure}[h!]
    \centering
    \includegraphics[width=0.8\textwidth]{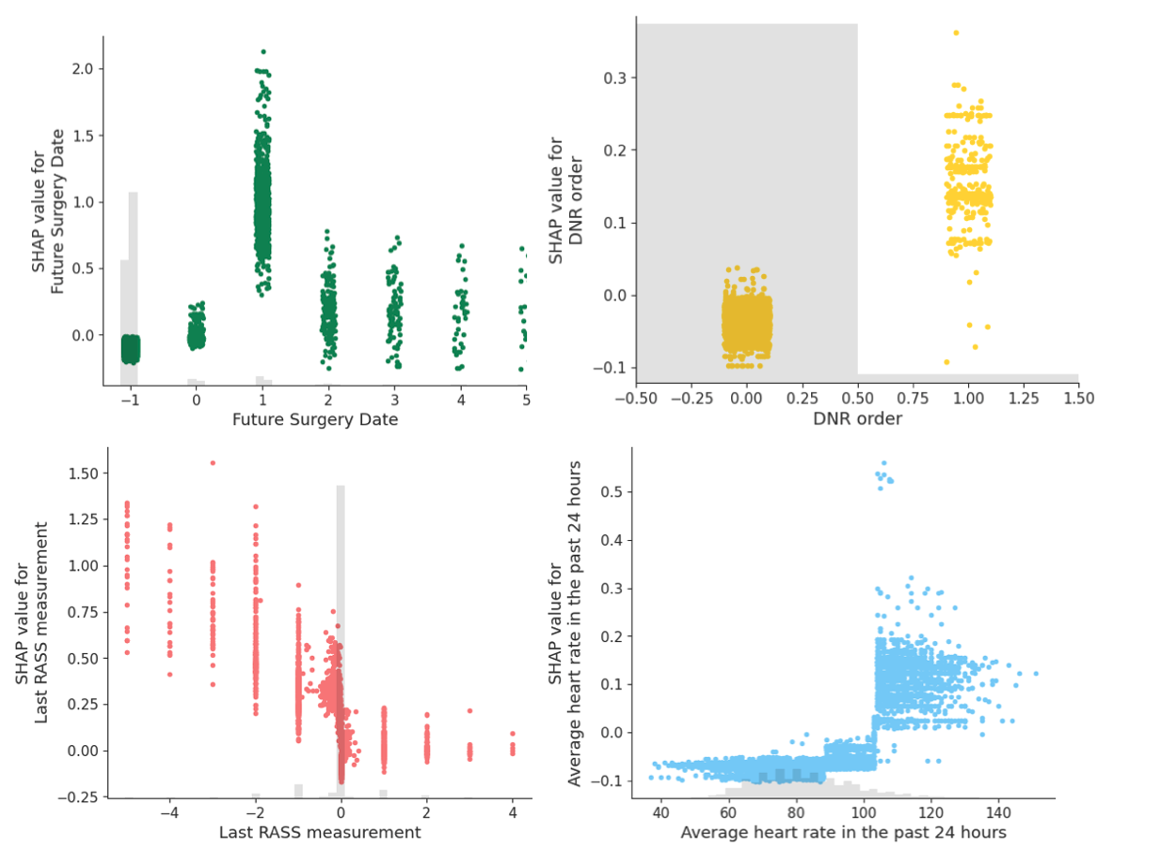} 
    \caption{SHAP interaction plots for future surgery date (ordered categorical), DNR order (binary categorical), last RASS measurement (ordered categorical), and average heart rate in the past 24 hours (continuous numerical). }
    \label{fig:fig3}
\end{figure}

We observe that each feature’s contribution to the prediction varies as their value changes in Figure 3, where each point corresponds to one sample in the test set. We choose features that are of continuous numerical, ordered categorical, and binary categorical nature. We observe that a positive future surgery date contributes positively to SHAP value, with its highest inflection point moving from negative to positive SHAP values occurring at 1. This implies that a scheduled next-day surgery, which could potentially fail or create further medical complications, increases near-term (24-hour) deterioration event risk. Similarly, placement of a DNR order (1), lower RASS measurement (less than 0), and average heart rate in the past 24 hours (above 105) all correspond to a positive SHAP value and a higher likelihood of a deterioration event occurrence. Note that the trend of SHAP value for features could be both linear (negative last RASS measurement) and nonlinear (future surgery date), demonstrating the ability of SHAP to capture rich feature relationships. 

\section{Deployment in Production}
\subsection{Implementation Considerations}
Several critical factors were considered during the development of the predictive model to ensure its efficiency and applicability in real clinical settings. Firstly, to ensure scalability for rapid prediction updates, the model is designed to be lightweight, enabling rapid processing of individual samples. To ensure data integrity, significant amounts of checks were implemented to ensure that recording delays, erroneous data entries, as well as data leakage would not negatively impact the data quality. These checks include the cleaning of data entries with different structures, excluding samples outside of the expected time range of admission, as well as ensuring that no recorded data directly corresponds to the prediction of outcome. Given the high-stakes nature of the environments in which the model would be deployed, another crucial aspect was the development of straightforward, direct visualizations. These visualizations are tailored to enable physicians to utilize the tool optimally under time pressure. Consequently, the selection of threshold cutoffs for alerts was extensively deliberated in collaboration with physicians, as also previously discussed in Section 3.3. This collaboration ensured that the binary alerts generated by the model were both clinically valid and reasonable, thereby supporting swift and accurate decision-making in critical care scenarios. Finally, model calibration was evaluated to ensure that the model output probability corresponds to the empirical event rate, as can be seen in Appendix Figure 1. 

\begin{figure}[h!]
    \centering
    \includegraphics[width=\textwidth]{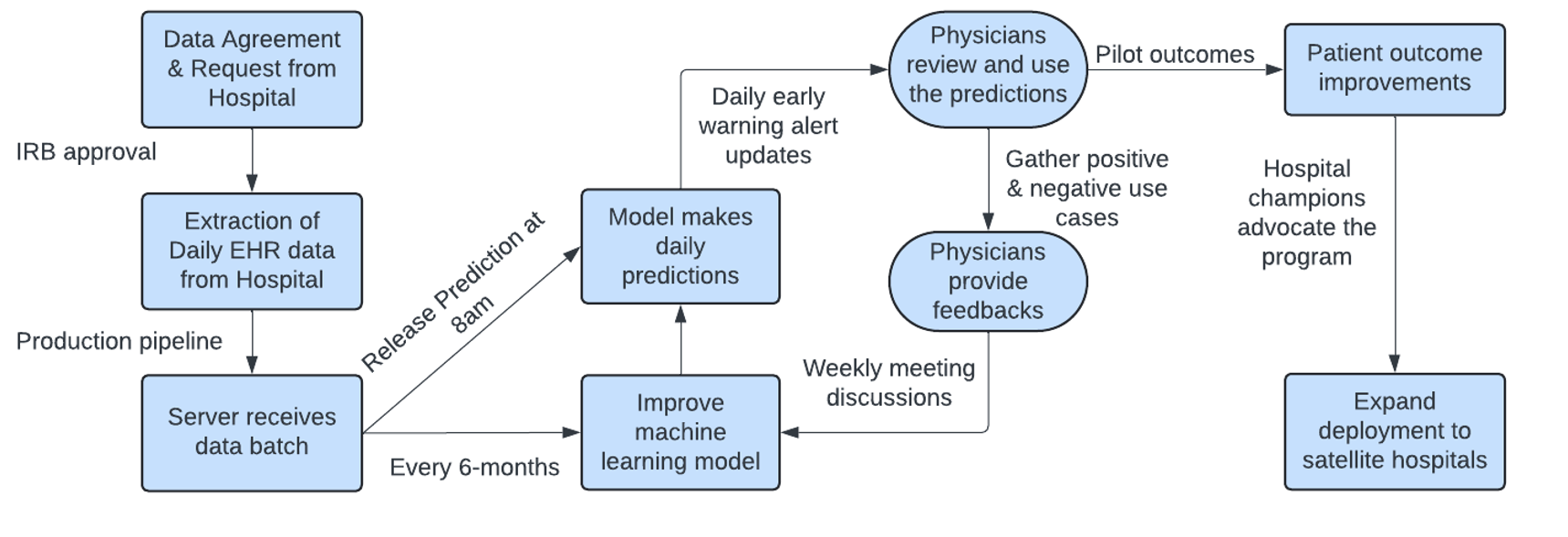} 
    \caption{The general pipeline from data extraction and model development to physician adoption for our existing implemented early warning alert in a large U.S. hospital.}
    \label{fig:fig4}
\end{figure}

\subsection{Currently Implemented Pipeline}
The proposed model has since been implemented and is currently under testing in the hospital system with daily updated data, which is received at 11:59 pm every day. We demonstrate the implementation pipeline in Figure 4, which largely follows the one in [18] for patient inflow predictions. The pipeline is developed by a proprietary firm that processes daily updated patient information received from the hospital system, which is used to make early warning alerts for each patient at 8am in the morning prior to the first patient status update round. These alerts are reviewed by the physicians, who gather both positive and negative use cases, and report back to the model maintenance team to investigate and update the model. However, the model’s performance could deteriorate as new data comes in and the patient covariate distribution experiences distributional shifts. To ensure model performance do not deteriorate due to distributional shift, after the first round of model training, the model is kept static until being retrained again every 6 months. As we improve patient outcomes, we expand the model deployment to satellite hospitals of the same hospital system and then beyond to other hospital systems.

\subsection{Forecast Dashboard}
We designed and implemented the following dashboard in the participating hospital and daily process around 1500 patients across the seven satellite locations. For each patient, the dashboard displays department name, room number, bed label, and current service to facilitate quick and accurate patient location identification by caregivers. Computed risks of deterioration as well as changes of deterioration risks in comparison to the previous one and two days are also displayed. Red alerts indicate the highest-risk patient group, where the patient’s current day (the immediate next 24 hour) deterioration risk is more than 12\%, or has increased over the previous day by 6\%; yellow alerts indicate the medium-risk patient group, where the patient’s current day deterioration risk is more than 3\%, or has increased over the previous day by 1.5\%; and white alerts indicates the low-risk patient group for all remaining patients. These risk stratifications are used as triage tools for the physicians during rounding to avoid delays in identifying and prioritizing patients who need check-ins and interventions most urgently. 

\begin{figure}[h!]
    \centering
    \includegraphics[width=\textwidth]{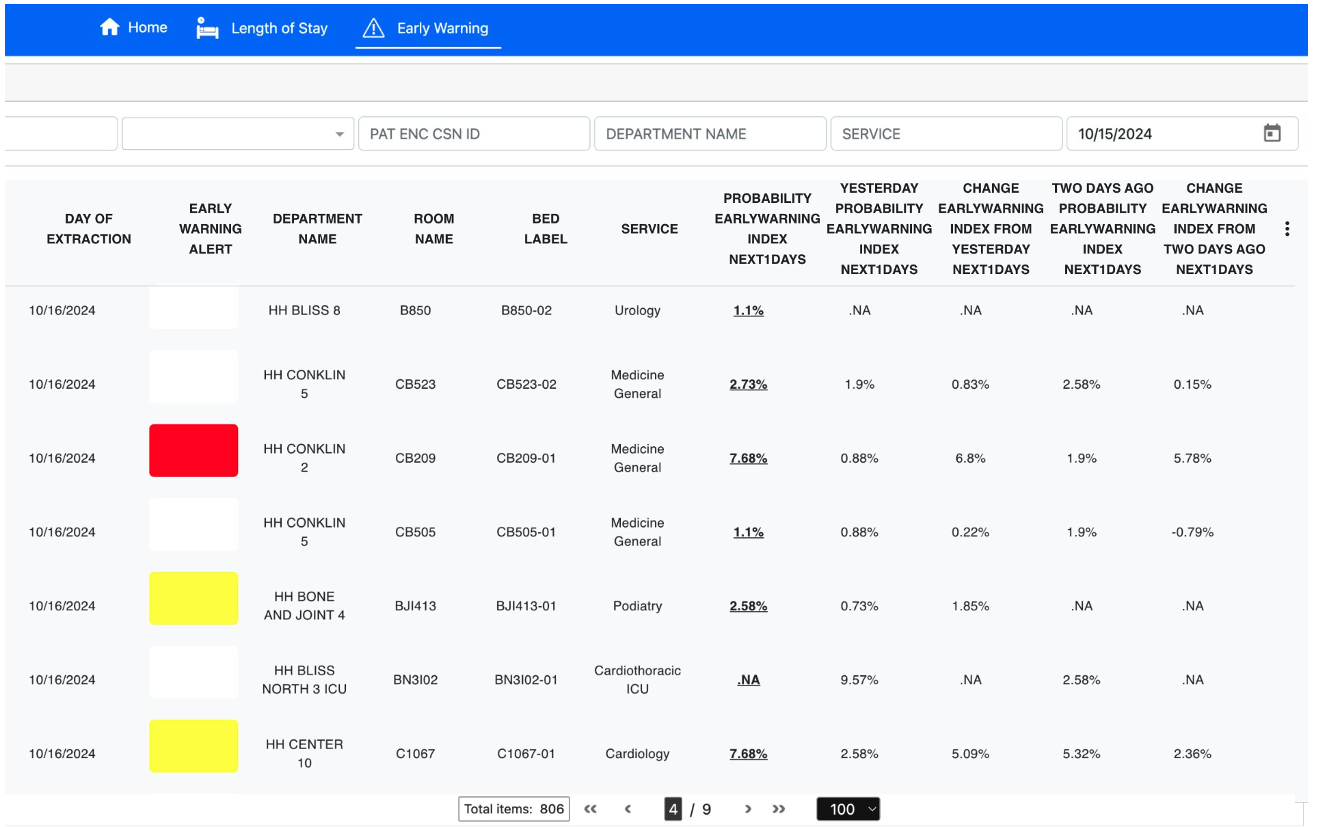} 
    \caption{Dashboard currently implemented at participating hospital. }
    \label{fig:fig5}
\end{figure}

\section{Managerial Implications and Limitations}
\subsection{Operational Benefits via Early Warning Alerts}
The implementation of early warning systems for patient deterioration is critical and addresses the complex, multifaceted dynamics of planning and preparation across departments of hospital facilities. The primary goal of these systems is to provide clinicians with advanced notice, allowing them to intervene before deterioration occurs. Once a warning is triggered, specific conditions—such as abnormal blood measurements or signs of respiratory collapse—can be promptly assessed to allow rapid identification of the root cause of a patient's decline. Physicians can then take targeted actions to stabilize the patient by various clinical interventions  Targeted actions can also include change in the level of care by moving the patient to ICU or step down for better monitoring and advanced interventions. By offering early intervention opportunities, these systems not only prevent the need for critical resuscitation efforts but also reduce the physical toll on patients and alleviate the significant burden on healthcare resources.

Physician burnout is another critical issue that could be improved by the integration of such an early warning system. Specifically, these burnouts are closely tied to the unpredictability of inpatient settings, where fluctuating patient volumes and clinical conditions make it difficult for physicians to plan their shifts efficiently. This unpredictability often leads to overtime, contributing to low workplace satisfaction and increased physician turnover. Upon starting rounds, physicians may have 20 or more patients to assess, with no real-time, objective method for determining which patients require immediate attention. Our model addresses this challenge by streamlining rounding activities and providing physicians with a quantified, objective tool to prioritize patient care. This "snapshot" prioritization enables quick identification of high-risk cases, reducing the need for manual reviews of vital signs and nurse notes. For low-risk patients, the tool can also aid in assessing discharge readiness—a key factor in downstream bed management and overall hospital logistics. By improving response times and decision-making accuracy, this system can help physicians plan their day more efficiently and ultimately reduce the strain that leads to burnout.

Our model is planned to continuously adapt to the dynamic nature of patient admissions by updating every 24 hours, ensuring that predictions remain aligned with institutional shifts and reflect the most current patient data. Once fully implemented, the model will support approximately 40-50 medical providers and 100-120 nurses across various departments, with the Department of Medicine alone managing 370-400 patients. In the long term, the model’s reach is anticipated to expand across seven hospitals within the same healthcare system, offering comprehensive support to preempt critical care interventions, reduce unnecessary physician burnout, and ultimately lead to proactive patient management.

\subsection{Patient Outcome Impacts}
Early warning alerts for patient deterioration facilitate timely and precise interventions, preventing the progression of conditions that could otherwise escalate into severe complications. This capability not only improves the immediate responsiveness of healthcare teams but also reduces the need for more intensive and costly treatments in the future. By intercepting deterioration at an early stage, these systems could help maintain patient stability and prevent acute crises that typically result in high-cost care. Concrete outcomes of implementing such alert systems include measurable reductions in the length of hospital stays and ICU admission rates. Shorter hospital durations are directly correlated with reduced healthcare costs and lower patient exposure to hospital-acquired infections, which are significant contributors to morbidity. Furthermore, early stabilization of patient conditions decreases the risk of readmissions, a key indicator of quality care and an important metric for hospital performance evaluations.

\subsection{Limitations}
We note several limitations of our study and propose potential solutions to mitigate them in future studies. First, the exclusion of patients currently in the Intensive Care Unit (ICU) from the cohort means that the model's predictive capabilities do not extend to the subset of the hospital population that is already under critical conditions. This omission restricts the model's effectiveness in predicting emergency responses and mortality among the most critically ill patients, who could potentially benefit the most from accurate early mortality warning systems. 

Secondly, the study could benefit from incorporating data from other modalities, such as medical images or clinical notes. The inclusion of imaging data, for example, could offer critical insights into conditions that are not obviously measured in vital signs or lab results. Similarly, although we make use of LLMs, the availability of additional clinical notes, such as nurse notes and radiology notes, could provide additional valuable information such as psychological issues, previous admission history, or other more rarely recorded information. The choice to opt out of these modalities in the current study is due to the consideration that both notes and image data are well known for having processing delays in data recording systems, which would not be readily available often for real-time alarms. In addition, clinician and other related notes contain large amounts of personally identifiable information that would require careful removal process before being able to be used by third-party data storage or process service, which often involve significant manual labors and physician reviews. 

We also note that the operational effects of the deterioration index cannot be straightforwardly evaluated. This is largely due to the presence of many endogenous factors that are out of the control of any single model predictions but rather a combination of logistical factors within the hospital organizational structure. However, although these changes are highly subjective, by providing this clinical support tool to the physicians, inefficiencies regarding data processing and integration of multimodal information can be alleviated and provide value to hospital managers. 

Lastly, we note that the existing system was developed for a 24-hour continuous update schedule, however, physician and nurse schedules could be shorter than this existing timeline. An updated release of 6-hour or 8-hour interval predictions could potentially align more closely with their shift changes.

% Acknowledgments here
\ACKNOWLEDGMENT{%
\textbf{Author Contributions}
Y.M. led the efforts of model development, planning and performing experiments, writing code, analyzing results, and writing the manuscript, and contributed to data processing and model implementation. K.V.C. contributed to data processing, developed models, planned and performed experiments, wrote code, analyzed results, and edited the manuscript. G.S. contributed to the research, validation of the results, and edited the manuscript. J.M. collected data used in the study. M.L. wrote code, contributed to model development and analyzing results, and implementation of the model in the hospital system. H.H. supervised 
and led the medical team, contributed to experimental design and model evaluation, and edited the manuscript. D.B. directed the overall project, from concept and research to implementation, and edited the manuscript.

\noindent
\textbf{Acknowledgments}
The authors would like to thank the Hartford HealthCare team for their help with implementation, feedback, and discussions. We are grateful to Jeff Mather from Hartford and Ali Haddad-Sisakht for their support in data extraction. The authors Anne Gvozdjak, Aaron Y. Zhu, and Demetrios C Kriezis for providing support on computational experiments to our work. Finally, we acknowledge the MIT SuperCloud and Lincoln Laboratory S0upercomputing Center for providing computing resources and technical consultation 
}% Leave this (end of acknowledgment)

% Appendix here
% Options are (1) APPENDIX (with or without general title) or 
%             (2) APPENDICES (if it has more than one unrelated sections)
% Outcomment the appropriate case if necessary
%
% \begin{APPENDIX}{<Title of the Appendix>}
% \end{APPENDIX}
%
%   or 
%
% \begin{APPENDICES}
% \section{<Title of Section A>}
% \section{<Title of Section B>}
% etc
% \end{APPENDICES}

% References here (outcomment the appropriate case) 

% CASE 1: BiBTeX used to constantly update the references 
%   (while the paper is being written).
\newpage
\bibliographystyle{informs2014} % outcomment this and next line in Case 1
\bibliography{ref.bib} % if more than one, comma separated

% CASE 2: BiBTeX used to generate mypaper.bbl (to be further fine tuned)
%\input{mypaper.bbl} % outcomment this line in Case 2

\end{document}